\documentclass[a4paper, 10pt, conference]{ieeeconf}      
\usepackage{FG2020}

\FGfinalcopy 

\IEEEoverridecommandlockouts                              
\usepackage{graphics} 
\usepackage{graphicx}
\usepackage{subcaption}
\usepackage{epsfig} 
\usepackage{mathptmx} 
\usepackage{times} 
\usepackage{amsmath} 
\usepackage{amssymb}  
\usepackage{diagbox}  
\usepackage{multirow} 
\usepackage{booktabs} 
\usepackage{xcolor}
\usepackage{algorithm}
\usepackage[noend]{algpseudocode}
\usepackage{multirow}
\usepackage{makecell}
\usepackage{helvet} 
\usepackage{courier}  
\usepackage[hyphens]{url}  
\usepackage{cite}

\def\FGPaperID{****} 

\title{\LARGE \bf
Multitask Emotion Recognition with Incomplete Labels
}

\author{\parbox{16cm}{\centering
    {\large Didan Deng, Zhaokang Chen, Bertram E. Shi}\\
    {\normalsize
    Department of Electronic and Computer Engineering, \\Hong Kong University of Science and Technology, Kowloon, Hong Kong\\
    \{ddeng,zchenbc\}@connect.ust.hk\qquad\quad eebert@ust.hk}}
}

\begin{document}

\ifFGfinal
\thispagestyle{empty}
\pagestyle{empty}
\else
\author{Anonymous FG2020 submission\\ Paper ID \FGPaperID \\}
\pagestyle{plain}
\fi
\maketitle

\begin{abstract}

We train a unified model to perform three tasks: facial action unit detection, expression classification, and valence-arousal estimation. We address two main challenges of learning the three tasks. First, most existing datasets are highly imbalanced. Second, most existing datasets do not contain labels for all three tasks. To tackle the first challenge, we apply data balancing techniques to experimental datasets. To tackle the second challenge, we propose an algorithm for the multitask model to learn from missing (incomplete) labels. This algorithm has two steps. We first train a teacher model to perform all three tasks, where each instance is trained by the ground truth label of its corresponding task. Secondly, we refer to the outputs of the teacher model as the soft labels. We use the soft labels and the ground truth to train the student model. We find that most of the student models outperform their teacher model on all the three tasks. Finally, we use model ensembling to boost performance further on the three tasks. Our code is publicly available\footnote{https://github.com/wtomin/multitask-Emotion-Recognition-with-Incomplete-Labels}.

\end{abstract}

\section{INTRODUCTION}
Video emotion recognition is a longstanding problem studied by computer scientists and psychiatrists. It seeks to recognize people's emotional state automatically based on videos of their behaviour. Several taxonomies have been proposed to quantify human emotion. Facial Action Unit coding system was proposed by Ekman  and  Friesen~\cite{friesen1978facial}. An action unit (AU) is a fundamental action of an individual muscle or group of muscles. Although the facial AU coding system can describe instantaneous changes in expression, it is quite hard to understand for non-experts. In the contrast, the seven basic emotions proposed by Ekman and  Friesen~\cite{ekman1971constants} (anger, disgust, fear, happiness, sadness, surprise and neutral) are much easier to understand. Gunes and Pantic~\cite{gunes2010automatic} proposed the continuous two-dimensional Valence-Arousal system. Valence describes how positive or negative the emotion is. Arousal describes how active or calm the person is. Most existing emotion datasets include only one set of labels~\cite{goodfellow2013challenges,dhall2015video,barros2018omg}. Very few datasets~\cite{benitez2017emotionet} contain two or more sets of labels, due to the high cost and time required to annotate. Due to these limitations, most past work in emotion recognition has focused on only one type of label e.g. ~\cite{deng2019mimamo} for valence-arousal.

\begin{figure}[tb] 
\centering
\resizebox{1\columnwidth}!{
\includegraphics[width=\columnwidth]{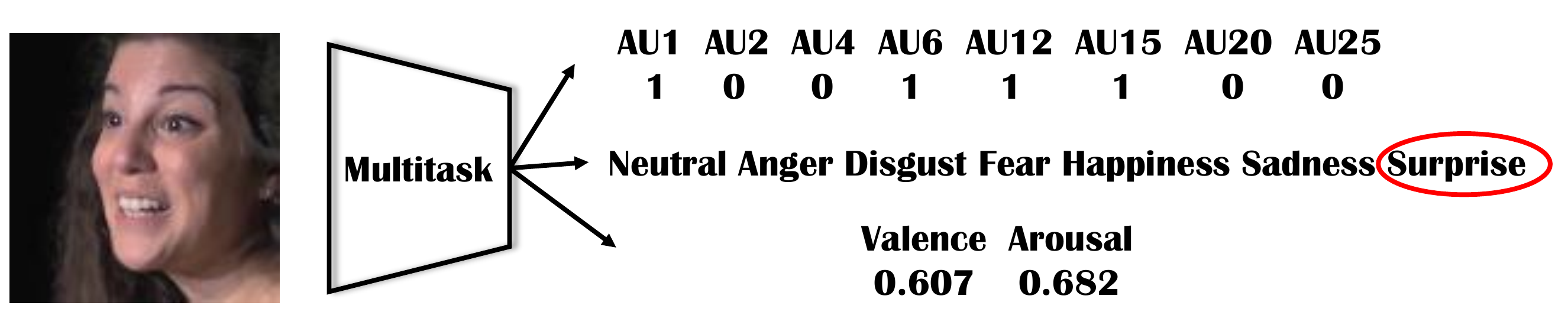} 
}
\caption{The multitask system for simultaneous frame-by-frame facial action unit detection, expression classification, and valence-arousal estimation. Most emotion datasets contain only one type of emotion label.}
\label{fig: teaser}
\end{figure}

A multitask system that can label an input with all sets of labels is an important goal. Such a system would be more efficient, and would meet the demands of a wide range of applications. As Figure~\ref{fig: teaser} shows, given a input facial video, the multitask system described here needs to detect the presence or absence of eight action units (AUs), classify the emotion, and estimate the valence-arousal values, in each frame. In this paper, we address two key challenges of multitask emotion recognition: data imbalance and missing labels.

The problem of imbalanced data is very common in both single-label and multi-label emotion datasets. For example, the FER2013 dataset \cite{goodfellow2013challenges} consists of 35,887 facial images, each annotated with one of seven basic emotions. However, about $25\%$ of the images are labeled with "happy", while only $1.5\%$ are labeled with "disgust". The dramatic difference between the numbers of samples in the majority and minority classes leads to an overemphasis on the majority class, which hinders overall performance. We propose to deal with data imbalance by combining two methods: the introduction of more data, and re-sampling that over-samples minority classes and under-samples majority classes.

The problem of missing labels arises in multitask learning because most datasets are labelled for only one task.  An intuitive solution is Binary Relevance (BR), which trains one classifier for each class in each task. However, BR fails to model the correlations between different labels. It is not efficient to train many classifiers, especially when the number of classes increases. To overcome these shortcomings, we propose to use a shared feature extractor for all tasks, and to use multiple heads on top of the feature extractor as classifiers. To better learn inter-task correlations with incomplete labels, we first train a teacher model with missing labels, and then use the ground truth and the outputs of the teacher model as supervision for a set of student models. Teacher-student networks are commonly used in knowledge distillation~\cite{hinton2015distilling}, which seeks to to compress model size and reduce inference time. However, in our case, the student model has the same structure and size as the teacher model. We hypothesize that a student exposed to a complete set of imperfect labels will learn better than a teacher exposed to an incomplete set of ground truth labels.

Our primary contributions are:
\begin{quote}
    \begin{itemize}
        \item We highlight the importance of data balancing for classification tasks in multitask learning. Surprisingly, we find that data balancing is not always beneficial for regression tasks.
        \item We propose an to learn from missing labels using one teacher model and several student models. This algorithm is generic in the sense that we do not assume any particular structure (e.g. CNN or RNN) for the teacher/student networks,
    \end{itemize}
\end{quote}

\section{Related Work}
\subsection{Data Balancing}
The data imbalance problem is very common in emotion datasets~\cite{goodfellow2013challenges,kossaifi2017afew,zafeiriou2017aff,kollias2019deep,kollias2017recognition}. 
Many methods have been proposed to solve the data imbalance problem. As a first priority, we should collect more data if possible. Where this is not possible, we can resample the data or generate synthetic samples. Resampling has been proved effective in dealing with imbalanced datasets~\cite{garcia2012effectiveness}. Since it is classifier-independent, it can be easily applied to many applications. Charte et al.~\cite{charte2015addressing} proposed a number of measures of the imbalance in multilabel datasets, such as the MeanIR (Mean Imbalance Ratio), as well as rebalancing algorithms for multilabel datasets.

Some work has focused on algorithm adaptation methods, such as the cost-sensitive learning~\cite{sun2007cost}. The cost-sensitive SVM~\cite{cao2013optimized} incorporates the evaluation metrics (AUC and G-mean) into the objective function of SVM. The example-dependent cost-sensitive decision tree algorithm~\cite{bahnsen2015example} takes example-dependent costs into account when training and pruning a tree. Such algorithm-specific modifications are useful in practice, but not flexible enough to be applied to a broader range of applications. 

\subsection{Learning from Missing Labels}

To learn from missing labels, the most intuitive approach is to learn one classifier for each class, which is called as Binary Relevance (BR)~\cite{luaces2012binary}. Another method replaces missing labels with negative labels~\cite{sun2010multi,bucak2011multi}. Although simple, this method impedes the model performance because many invisible positive ground truth labels are set to negative labels. There are some assumption-based methods to complete the training labels. For example, based on a low rank label matrix assumption, Cabral et  al.~\cite{cabral2011matrix} used matrix completion to fill in the missing labels. Based on the assumption that the missing labels are latent variables, Kapoor et al.~\cite{kapoor2012multilabel} used Bayesian networks to infer missing labels. More specifically in emotion recognition, Kollias et al.~\cite{kollias2019face,kollias2019expression} used a rule-based method to complete the training labels based on the co-occurrence between expression categories and AUs. Our method makes no hypotheses on the underlying relations over tasks. Instead, we use a data-driven teacher model to fill in the missing labels.

\subsection{Knowledge Distillation}

Hinton et al.~\cite{hinton2015distilling} proposed Knowledge Distillation for model compression. The knowledge of a larger network is transferred to a relatively smaller network using a modified cross entropy loss function. They introduce a new hyper-parameter called \textit{temperature} $T$ into the softmax function, and suggest that setting $T>1$ can increase the weight of smaller logit values, thus providing dark knowledge. In other words, the relative probabilities can reveal more information about inter-class relations than the one-hot labels. Knowledge Distillation has been proved effective in model compression, continual learning~\cite{li2017learning} and domain adaptation~\cite{asami2017domain}. However, its application to multitask learning with missing labels is under-researched.

Knowledge distillation for regression is not as common as for classification. Some work in face alignment~\cite{lee2018teacher,wang2017model} has used the  L1 or L2 distance as the distillation loss function. To enable the use of knowledge distillation using the hyper-parameter \textit{temperature} for valence-arousal estimation, we transform the regression task to a classification task by discretizing the continuous values. Then we can use \textit{temperature} to control the smoothness of the soft labels.

\begin{figure*}[ht]
  \centering
  \begin{subfigure}[b]{0.45\textwidth}
    \includegraphics[width=\textwidth]{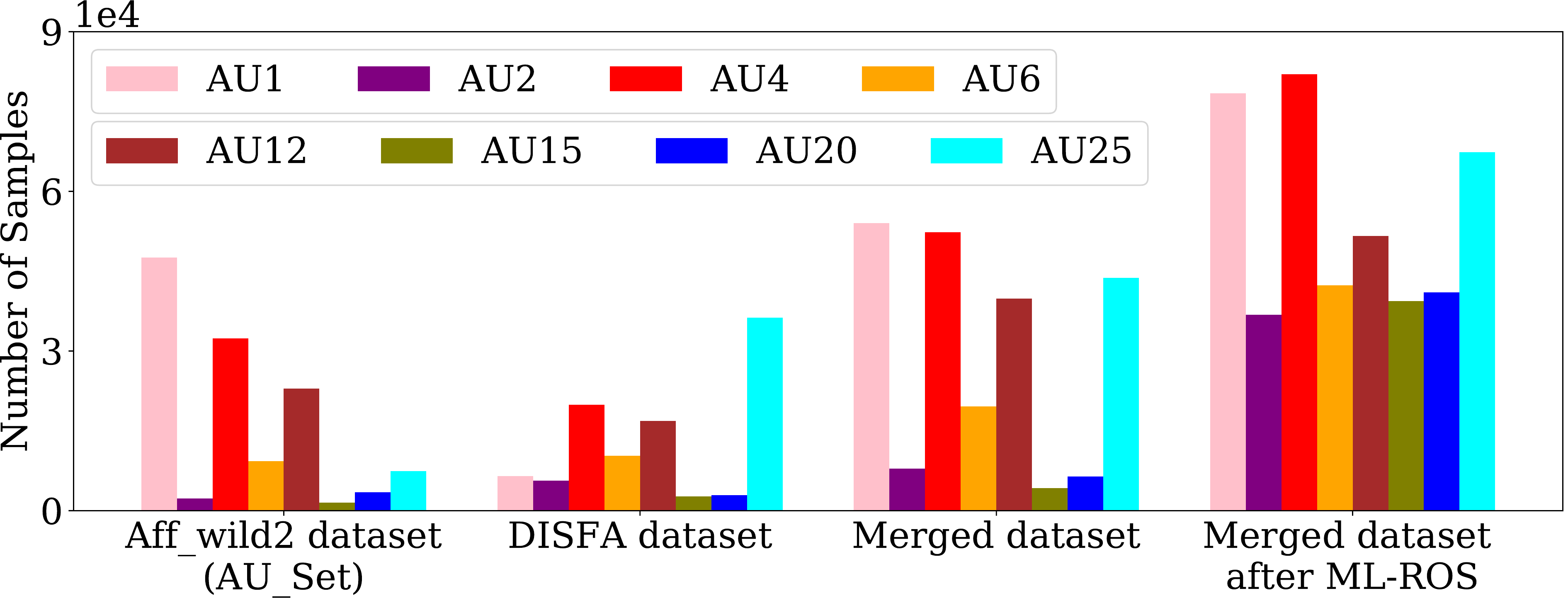}
    \caption{AU distributions.}
  \end{subfigure}
  \begin{subfigure}[b]{0.45\textwidth}
    \includegraphics[width=\textwidth]{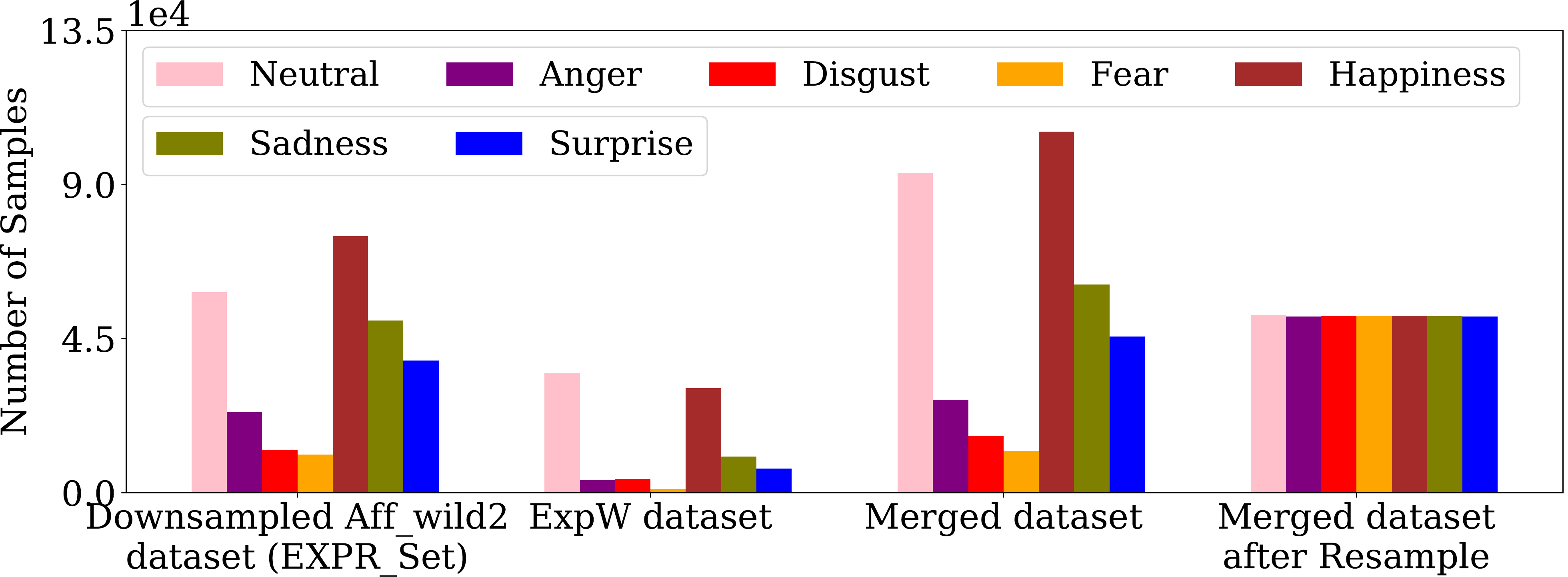}
    \caption{Facial Expression distributions.}
  \end{subfigure}
    \begin{subfigure}[b]{0.9\textwidth}
    \includegraphics[width=\textwidth]{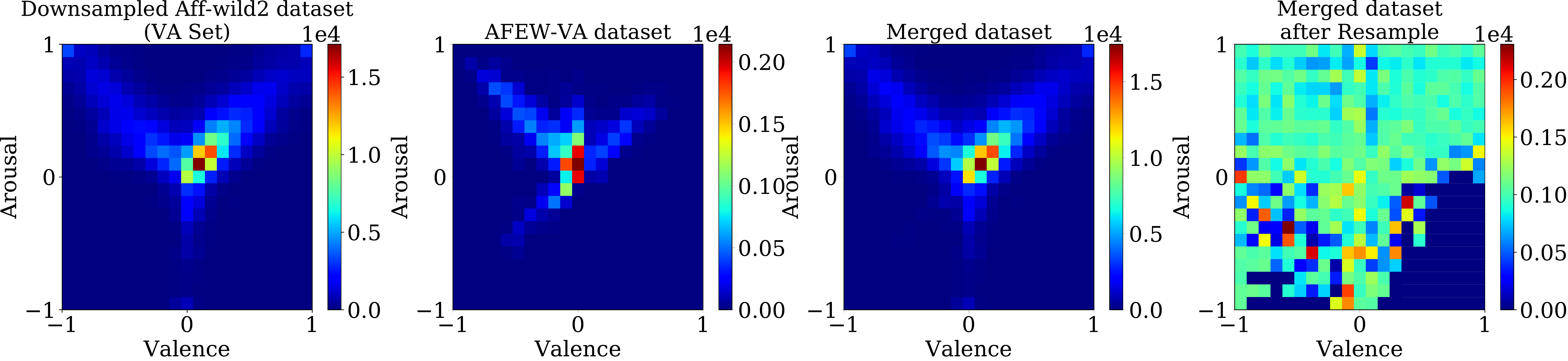}
    \caption{Valence and Arousal distributions.}
  \end{subfigure}
  \caption{Comparison of different distributions.}
  \label{fig:dis}
\end{figure*}

\section{Methodology}

In this paper, we design a network to perform three tasks simultaneously: facial AU detection, expression classification, and valence-arousal estimation. Facial AU detection is a multi-label classification problem, where the model detects the presence/absence of eight AUs (AU1, AU2, AU4, AU6, AU12, AU15, AU20, AU25), which are not mutually exclusive. Expression classification maps each frame to one of seven basic emotions. Valence-arousal estimation is a regression problem, where the model estimates two continuous scores in the interval $[-1, 1]$.

\subsection{Data Balancing}
The main dataset that we use is the Aff-wild2 dataset~\cite{kollias2018aff}, which is an in-the-wild video dataset. There are three sub-datasets in the Aff-wild2 dataset: one for facial AU detection, one for expression classification, and the other for valence-arousal estimation. Each sub-dataset contains several videos, in which every frame is annotated with related labels. More details about this dataset can be found in~\cite{kollias2020analysing}.

We address the data imbalance problem in the Aff-wild2 dataset by  importing external datasets followed by rebalancing.

For facial AU, we import the Denver Intensity of Spontaneous Facial Action (DISFA) dataset~\cite{mavadati2013disfa}. By merging the DISFA dataset with the Aff-wild2 dataset, we enlarge the data size and also increase the number of samples in the minority classes. However, simply merging the two datasets does not solve the data imbalance problem. We use the ML-ROS algorithm~\cite{charte2015addressing} to oversample instances with positive minority labels. After applying ML-ROS, we find that the numbers of samples for each AU become closer. The facial AU distributions of the Aff-wild2 dataset, the DISFA dataset and the merged dataset are shown in Figure~\ref{fig:dis} (a). As it shows, the distribution of the dataset after ML-ROS is not fully balanced, which is caused by the co-occurrence of some minority labels and majority labels in the same instances.

For expression classification, we import the Expression in-the-Wild (ExpW) Dataset~\cite{zhang2018facial}, which contains 91,795 images annotated with seven emotion categories. Since the number of images in the expression set of the Aff-wild2 dataset is 10 times larger than in the ExpW dataset, we downsample the Aff-wild2 dataset to make their sizes comparable. After merging the downsampled Aff-wild2 dataset and the ExpW dataset, we resample the samples to ensure the instances of each class have the same probability of appearing in one epoch. The distributions of seven emotion categories in the downsampled Aff-wild2 dataset, the ExpW dataset and the merged dataset are shown in Figure~\ref{fig:dis} (b).

For valence-arousal estimation, we import the AFEW-VA dataset~\cite{kossaifi2017afew} which contains 30,051 frames annotated with both valence-arousal scores in $[-10, 10]$, which are rescaled to $[-1, 1]$. We downsample the Aff-wild2 dataset by 5, and merge it with the AFEW-VA dataset. We then discretize the valence-arousal scores into 20 bins of the same width. We treat each bin as a category, and apply the oversampling/undersampling strategy as used in the expression recognition task. The distributions of the valence-arousal scores in the downsampled Aff-wild2 dataset, the AFEW-VA dataset and the merged dataset are shown in Figure~\ref{fig:dis} (c). The merged dataset is resampled to improve balance, but because of some rare cases, e.g., $(V,A) = (1, -1)$, the distribution of the resampled dataset is not fully balanced.

\begin{figure*}[ht] 
\centering
\includegraphics[width=0.8\textwidth]{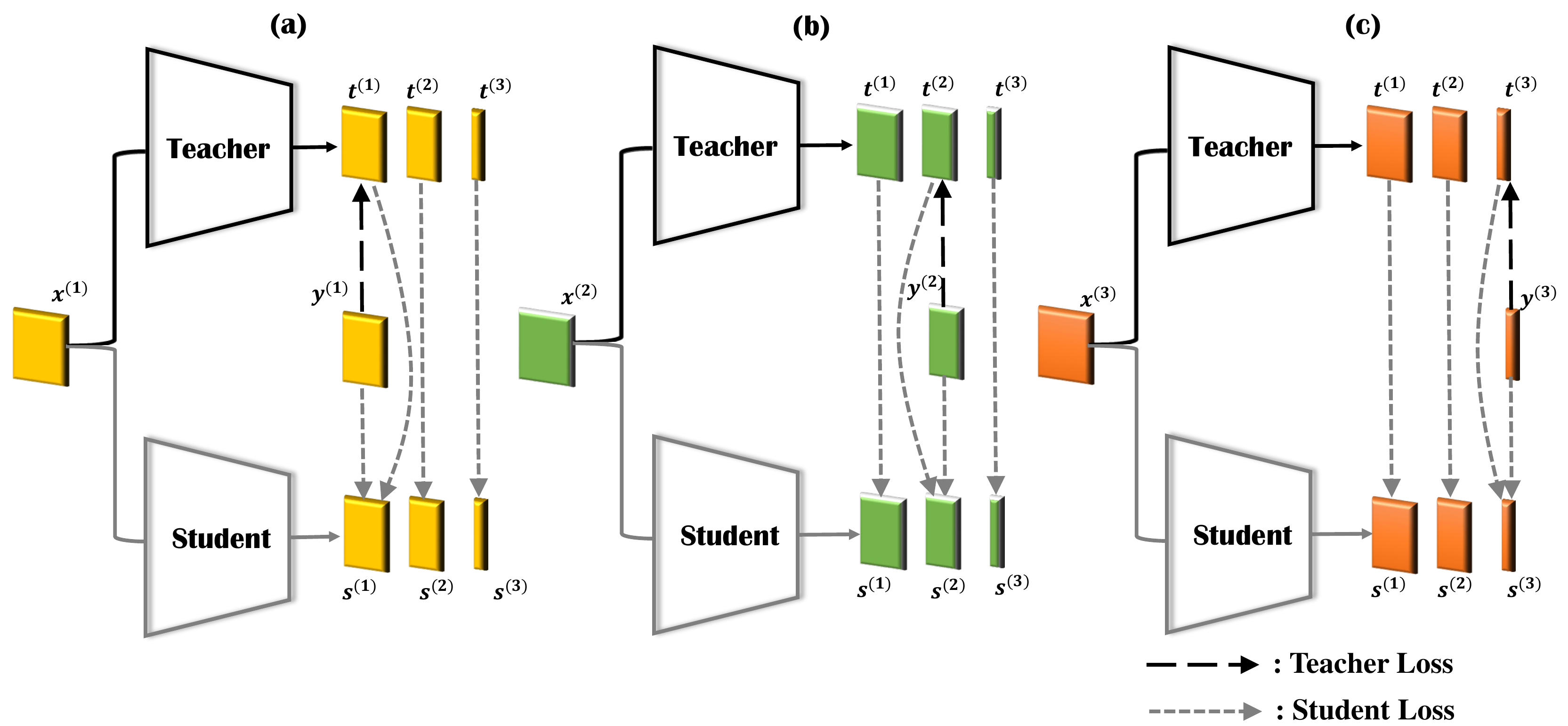} 
\caption{Diagram of Algorithm \ref{algorithm1}. First, we train the teacher model with the teacher loss, which is shown in the top half of this figure. Then we train the student model with the student loss, which is shown in the bottom half of this figure. Although we show each task separately in (a), (b) and (c) for the ease of illustration, during training we include data from all the three tasks in each batch. }
\label{fig: algorithm}
\end{figure*}

\subsection{Learning from Missing Labels}

We denote the training dataset or a subset of the training dataset (i.e. a batch) by $(X,Y)$, where $X$ is a set of input vectors and $Y$ is a set of ground truth training labels. Although we wish to train a network to perform three tasks, we note that each data instance contains only a label for one task.  Thus the entire dataset or batch consists of three subsets, $(X,Y) = \{ (X^{(i)},Y^{(i)}) \}_{i=1}^{3}$. For convenience of notation, we assume each subset $i$ includes an equal number $N$ of instances within a batch, i.e. $(X^{(i)},Y^{(i)}) = \{ (x^{(i,n)},y^{(i,n)} ) \}_{n=1}^{N}$ where $n$ indexes the instance. The batch size is $3\cdot N$. It is straightforward to extend to the case where different subsets have different cardinality. For conciseness and simplicity of notation, we will often drop the indexing by instance, i.e refer to a label for task $i$ by $y^{(i)}$.

The inputs for all instances have the same dimensionality, independent of task.  However, the ground truth labels for different tasks have different dimensionality. The label for the first task (facial AU detection) is $ y^{(1)} \in \{ 0, 1 \}^{8}$. Similarly, $ y^{(2)} \in \{0,1\}^{7}$ (expression classification) and $y^{(3)} \in [-1,1]^2$ (valence-arousal estimation). 

As each instance only has the label from one task, the intuitive way to train a unified multitask model is to only use label from that task for supervision. However, this training strategy does not capture inter-task correlations. 

To capture these correlations, we propose the two-step algorithm shown in  Figure~\ref{fig: algorithm}. In the first step, we train a single teacher model using only ground truth labels. In the second step, we replace the missing labels with soft labels derived from the output of the teacher model. We then use the ground truth and soft labels to train multiple student models. 

We denote the output of a multitask network by $f_\theta(\cdot)$, where $\theta$ contains the model parameters, e.g. of the teacher or of the student network. Although each instance in the training data only has a label from one task, our model predicts the results for all the three tasks. We denote the output of the network for task $j$, by $f_\theta^{(j)} (\cdot)$. For example, $f_\theta^{(2)} (x^{(1)})$ indicates the output of the network for task 2 (expression estimation) for an instance in the training dataset that has only task 1 (facial AU detection) labels. To avoid clutter, we will often refer to the output of the teacher network on task $i$ by $t^{(i)}$ irrespective of what the input label is, i.e. $t^{(i)} = f_\theta^{(i)} (x^{(j)})$ for some $j \in \{1,2,3\}$ and similarly to the output of the student network on task $i$ by $s^{(i)}$, 

Generally speaking, the dimensionality of the teacher and student outputs for each task is the same as that of the ground truth. However, this is not true for the valence-arousal estimation task (task 3), where as described below, we convert the regression problem to a classification problem to facilitate distillation of teacher knowledge to the student.  

This architecture is quite generic, and can be applied to any network architecture placed inside the teacher and student blocks. Our main assumption is that the output layers of the teacher and student models are linear output layers. These can be converted into probabilities of each label or class using the element-wise logistic sigmoid $\sigma(\cdot)$ (e.g. for the facial AU detection task) or by a soft-max function for the multiclass classification. We define the soft-max function parameterized by temperature $T$ and applied to a $D$ dimensional vector $y = \{y_d\}_{d=1}^D$ by 
\begin{equation}
SF_d(y,T)= \frac{\exp(y_d/T)}{\sum_c \exp(y_c/T)} 
\label{eq:softmax}
\end{equation}

In training, the parameters of the teacher and student networks are obtained by minimizing different loss functions defined over each batch. These are are constructed by summing over different combinations of instance-wise loss functions which seek to bring the network output either closer to the ground truth, which we refer to as supervision loss functions, or closer to soft labels from the teacher, which we refer to as distillation loss functions. These are based on the binary cross entropy or cross entropy functions. For two vectors $y = \{y_d\}$ and $z = \{z_d\}$, we define the total binary cross entropy by
\begin{equation}
\mathrm{BCE}(y,z) = 
- \sum_d \{ y_d \cdot \log (z_d) 
+  (1- y_d) \cdot  \log(1-z_d)\}
\label{eq:BCE}
\end{equation}
and the cross entropy to be
\begin{equation}
\mathrm{CE}(y,z) = 
- \sum_d y_d \cdot \log (z_d)
\label{eq:CE}
\end{equation}

\subsubsection{Supervision loss functions}

We choose different loss functions for data coming from different tasks. 

Since facial AU detection (task 1) is a multilabel classification problem, we use the total binary cross entropy loss across all AUs, defined as follows:
\begin{equation}
\mathcal{L}^{(1)}(y^{(1)}, t^{(1)}) = 
\mathrm{BCE} \left( y^{(1)},\sigma(t^{(1)}) \right)
\label{eq:SL1}
\end{equation}

Since expression classification is a multi-class classification problem, we use the categorical cross entropy loss, defined as follows:
\begin{equation}
\mathcal{L}^{(2)}(y^{(2)}, t^{(2)}) = 
\mathrm{CE} \left(
y^{(2)}, \mathrm{SF}(t^{(2)},1)
\right)  
\label{eq:expr}
\end{equation}
where setting the temperature parameter to 1 in the parameterized soft-max function results in the standard soft-max function.

For valence-arousal estimation, we combine classification and regression losses. Although valence-arousal estimation is a regression problem, we transform it into a classification task by discretizing the range $[-1,1]$ into 20 bins and representing each scalar dimension (valence or arousal) as a 20 dimensional one-hot vector. We denote the function transforming a scalar continuous label to a 20 dimensional discrete label by $\mathrm{onehot}(\cdot)$. Thus, the categorical ground truth labels for valence and arousal are $\mathrm{onehot}(y_1^{(3)})$ and $\mathrm{onehot}(y_2^{(3)})$. Correspondingly, the output of the final FC layer for valence-arousal estimation, $t^{(3)}$ is 40 dimensional. The first 20 dimensions, $t_1^{(3)}$, corresponds to valence, and the second 20, $t_2^{(3)}$, corresponds to arousal.

The classification loss can be computed for each instance as:
\begin{equation}
\mathcal{L}_{\mathrm{class}}^{(3)} (y^{(3)}, t^{(3)}) =
\sum_{i=1}^2
\mathrm{CE} \left(
\mathrm{onehot}(y_i^{(3)}),\mathrm{SF} ( t_i^{(3)} , 1 )
\right)
\label{eq:va_class}
\end{equation}

To compute the regression loss, we first transform each 20 dimensional network output to a continuous scalar value by taking the dot product of the vector of bin centers $\vec{c}$ and the standard softmax applied to the output, i.e.
\begin{equation}
\Bar{t}_i^{(3)} = \vec{c} \cdot \mathrm{SF} ( t_i^{(3)} , 1 )  
\label{eq:d2c}
\end{equation}
The regression loss is computed over each batch, using the Concordance Correlation Coefficient (CCC) between the scalar outputs and the scalar ground truth labels,
\begin{equation}
CCC = \frac{2\rho_{y\bar{t}} \sigma_y \sigma_{\bar{t}}}{\sigma_y^2 + \sigma_{\bar{t}}^2 + (\mu_y - \mu_{\bar{t}})^2}
\label{eq:CCC}
\end{equation}
where $\rho_{y\bar{t}}$ is the correlation coefficient between the ground truth and the output and $\mu_y$, $\mu_{\bar{t}}$, $\sigma_y$, and $\sigma_{\bar{t}}$ are the means and standard deviations computed over the batch. We compute two CCC's, one for valence, $CCC_1$, and one for arousal $CCC_2$.

With a slight abuse of notation, we will define a per-instance loss function to be the sum of the $\mathcal{L}_{\mathrm{class}}^{(3)} (y^{(3)}, t^{(3)})$ and a portion of the negative CCC allocated to the instance. 
\begin{equation}
\mathcal{L}^{(3)}(y^{(3)},t^{(3)}) = \mathcal{L}_{\mathrm{class}}^{(3)} (y^{(3)}, t^{(3)}) + \frac{1}{B}\sum_{i=1}^{2}(1-CCC_i) 
\label{eq:va}
\end{equation}

\subsubsection{Distillation loss functions}

For AU detection, the distillation loss we use is the binary cross entropy loss between the teacher model outputs and the student model outputs:
\begin{equation}
\begin{split}
\mathcal{H}^{(1)}(t^{(1)}, s^{(1)}) =  \mathrm{BCE} \Big(\sigma(t^{(1)}), \sigma(s^{(1)}) \Big)
\end{split}
\label{eq:distill_fau}
\end{equation}
This is identical to the supervision loss in Eq.~(\ref{eq:SL1}) with the ground truth replaced by the teacher output, and the teacher by the student.

For expression classification, we define the distillation loss to be 
\begin{equation}
\mathcal{H}^{{(2)}}(t^{(2)}, s^{(2)}) = \text{CE}\Big(SF(t^{(2)}, T),SF(s^{(2)}, T)\Big)
\label{eq:distill_expr}
\end{equation}
This is very similar to the supervision loss in Eq.~(\ref{eq:expr}) with the ground truth replaced by the teacher output and the teacher by the student. The critical difference is that we set the temperature $T$ to be a value a greater than one when computing the teacher output. This further softens the teacher output in order to better reveal the dark knowledge contained in the teacher network. Based on a grid search, we set $T = 1.5$. 

For valence-arousal, we define the distillation loss to be \begin{equation}
\mathcal{H}^{{(3)}}(t^{(3)}, s^{(3)}) = \sum_{i=1}^2 \text{CE}\Big(SF(t_i^{(3)}, T),SF(s_i^{(3)}, T)\Big)
\label{eq:distill_va}
\end{equation}
As with the expression recognition distillation loss, this is very similar to the classification loss in Eq.~(\ref{eq:va_class}), with the critical difference being the increase in the temperature parameter, which we also set to $T = 1.5$.


\subsubsection{Batch-wise loss functions}

Given a batch of data $(X,Y) =  \{ \{ (x^{(i,n)},y^{(i,n)} ) \}_{n=1}^{N} \}_{i=1}^{3}$, we define different loss functions for training the teacher and student netorks. 

We denote the parameters of the teacher network by $\theta_\mathrm{t}$. The \textbf{teacher loss} is defined based only on supervision loss functions. 
\begin{equation}
\mathcal{F}_\mathrm{t}(X,Y,\theta_\mathrm{t}) = 
\sum_{i=1}^{3} \sum_{n=1}^{N}
\mathcal{L}^{(i)} \left(
    y^{(i,n)}, 
    f_{\theta_\mathrm{t}}^{(i)} ( x^{(i,n)} )
\right)
\label{eq:teacher loss}
\end{equation}
Since each instance only contains ground truth labels for one task, we use only the network output for that task and the corresponding supervised loss function in computing the loss function.

We denote the parameters of the student network by $\theta_s$. 
The \textbf{student loss} combines both supervision and distillation losses.
\begin{equation}
\begin{aligned}
\mathcal{F}_\mathrm{s}( & 
    X, Y, \theta_\mathrm{t},\theta_\mathrm{s}
) = 
\sum_{i=1}^{3} \sum_{n=1}^{N} 
\bigg\{ 
    \lambda \cdot
    \mathcal{L}^{(i)} \left(
        y^{(i,n)},
        f_{\theta_\mathrm{s}}^{(i)} ( x^{(i,n)} )
    \right) \\
    & + (1-\lambda) \cdot
    \mathcal{H}^{(i)} \left(
        f_{\theta_\mathrm{t}}^{(i)} ( x^{(i,n)} ),
        f_{\theta_\mathrm{s}}^{(i)} ( x^{(i,n)} )
    \right) \\
    & + \sum_{j \neq i} 
    \mathcal{H}^{(j)} \left(
        f_{\theta_\mathrm{t}}^{(j)} ( x^{(j,n)} ),
        f_{\theta_\mathrm{s}}^{(j)} ( x^{(j,n)} )
    \right)
\bigg\}
\label{eq:student}
\end{aligned}
\end{equation}
The parameter $\lambda$ determines how much we weight the ground truth versus the teacher on tasks where the ground truth exists for that data instance.  We set $\lambda = 0.6$ to weight the ground truth slightly more than the soft labels.

We note several key similarities and differences between the teacher (Eq. (\ref{eq:teacher loss})) and student  (Eq. (\ref{eq:student})) loss functions. First, we find that both functions exploit ground truth knowledge when it is available (line 1 in both equations). However, due to the $\lambda$ parameter, the student relies less on the ground truth. Instead, it also looks to the teacher for guidance (line 2 in Eq. (\ref{eq:student})). Recall, however, that the student does not seek to blindly follow the teacher, but rather a softened version of the teacher's output due to the increased temperature parameter in the distillation loss functions. Second, for each data instance, the teacher network primarily updates only those parameters associated with the labelled task, since the loss function is only computed on the outputs of the network for that task. Although there may be some sharing of information between tasks through the use of a shared feature extractor as we describe later, the final decisions on each task are based only on the ground truth labelled data of that task.  On the other hand, each instance contributes to the learning of all tasks by the student, both for the labelled task (lines 1 and 2 of Eq. (\ref{eq:student})) and for the unlabelled tasks (line 3 of Eq. (\ref{eq:student})) where the student relies upon the teacher for guidance.  However, as noted above, it is guided by a softer version of the teacher's output due to the increased temperature parameter. This increased temperature may better reveal dark knowledge, enabling the student model to generalize better than the teacher.

The pseudo code for training the teacher and the student models is shown as Algorithm~\ref{algorithm1}. Given a teacher model, we can repeat the student procedure in Algorithm~\ref{algorithm1} to obtain multiple student models.

    \begin{algorithm}
    \caption{Learning with Missing Labels}\label{euclid}
    \hspace*{\algorithmicindent} \textbf{Input} \\
    \hspace*{3em} dataset $D$ \\
    \hspace*{3em} parameters $\theta_\mathrm{t}$ (teacher),
    $\theta_\mathrm{s}$ (student) \\
    \hspace*{3em} epochs $N^e_t$ (teacher), $N^e_s$ (student)  \\ 
    \hspace*{3em} learning rate $\alpha$
    \begin{algorithmic}[1]
    \Procedure{Teacher}{}
    \State $\mathrm{i\_epoch} = 0$
    \While{$\mathrm{i\_epoch} < N^e_t$} 
    \While{$\mathrm{not\ epoch\ end}$} 
    \State $(X,Y) \gets \mathrm{sampler}(D)$
    \State $\mathrm{loss} = \mathcal{F}_\mathrm{t}
    (X,Y,\theta_\mathrm{t})$
    \Comment{Eq. (\ref{eq:teacher loss})}
    \State $\theta_\mathrm{t} \gets \theta_\mathrm{t} -\alpha \frac{\partial \mathrm{loss}}{\partial \theta_\mathrm{t}}$
    \EndWhile
    \State $\mathrm{i\_epoch} \gets \mathrm{i\_epoch} + 1$
    \EndWhile
    \EndProcedure
    \end{algorithmic}
    \begin{algorithmic}[1]
    \Procedure{Student}{}
    \State $\mathrm{i\_epoch} = 0$
    \While{$\mathrm{i\_epoch} < N^e_s$} 
    \While{$\mathrm{not\ epoch\ end}$} 
    \State $(X,Y) \gets \mathrm{sampler}(D)$
    \State $\mathrm{loss} = \mathcal{F}_\mathrm{s}
    (X,Y, \theta_\mathrm{t}, \theta_\mathrm{s})$ 
    \Comment{Eq. (\ref{eq:student})}
    \State $\theta_\mathrm{s} \gets \theta_\mathrm{s}
    -\alpha \frac{\partial \mathrm{loss}}{\partial \theta_\mathrm{s}}$
    \EndWhile
    \State $\mathrm{i\_epoch} \gets \mathrm{i\_epoch} + 1$
    \EndWhile
    \EndProcedure
    \end{algorithmic}
    \label{algorithm1}
    \end{algorithm}

\section{Experiments}
\begin{figure}[t] 
\centering
\resizebox{1\columnwidth}!{
\includegraphics[width=\columnwidth]{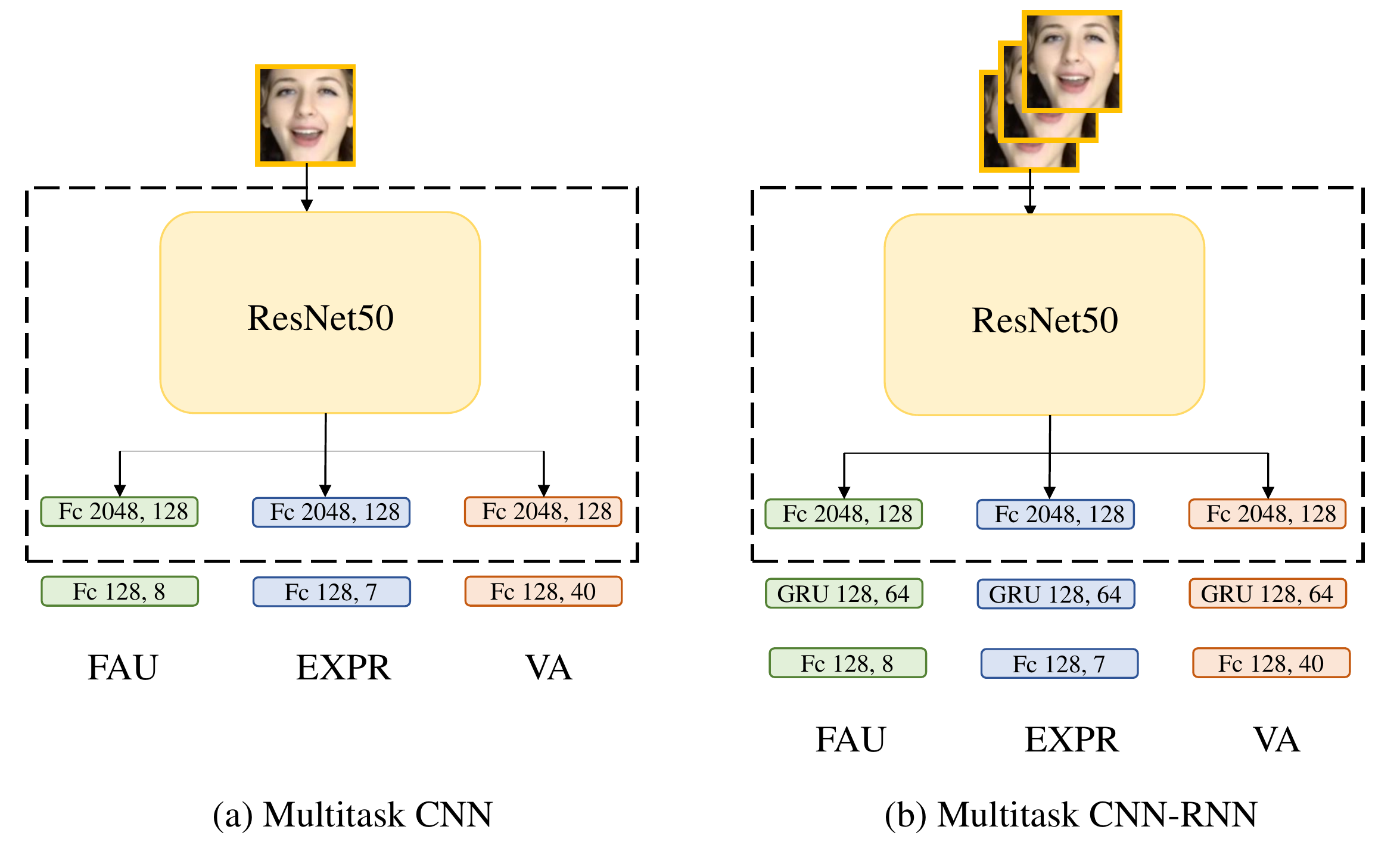} 
}
\caption{The multitask CNN (a) and CNN-RNN (b) architectures studied in the paper. The two architectures share the same ResNet spatial feature extractor shown in the dashed box. Parameters in this box were trained using the CNN architecture. }
\label{fig: models}
\end{figure}

We studied the two different network architectures shown in Figure~\ref{fig: models}: a CNN architecture and a CNN-RNN architecture. The CNN architecture considers each frame in isolation. It consists of a ResNet50 model, which functions as a shared feature extractor, followed by three MLPs stacked on top of the final ResNet50 conv layer: one for each task. The CNN-RNN architecture integrates information over time. It consists the same spatial feature extractor as that of the CNN architecture, followed by three bidirectional GRU layers to encode the temporal correlations over frames: one for each task.  The GRU layers are followed by fully connected layers generating the final outputs. The input image size to both architectures is $112 \times 112\times 3$.

We trained both single-task CNN architectures (three different ResNet50 networks each followed by a single MLP) and a multitask CNN architecture (one ResNet50 network followed by three MLPs). We use the single-task networks to aim to compare the performance of the teacher model with and without the addition of external data and with and without data balancing. Training of the  multitask architecture followed the procedures in Algorithm~\ref{algorithm1}. Both teacher and student models used the exact same architecture.  Multiple student models were ensembled.

 For training CNN-RNN model, we only used data from the Aff-wild2 dataset to train the parameters of the GRU and FC layers after the spatial feature extractor, because the external databases we used did not always contain image sequences. The parameters of the spatial feature extractor were the same as those learned by the multitask CNN architecture trained on both the Aff-wild2 and the external datasets. 
 
 We used Adam~\cite{kingma2014adam} to optimize both architectures. The learning rate of Adam optimizer was initialized to 0.0001, and decreased by a factor of 10 after every 3 epochs. The teacher model was trained for 8 epochs in total, the student model for 3. 

We used the same evaluation metrics as suggested in~\cite{kollias2020analysing}. For facial AU detection, the evaluation metric was $0.5 \cdot \mathrm{F1} + 0.5 \cdot \mathrm{Acc}$, where F1 denotes the unweighted F1 score for all 8 AUs, and Acc denotes the total accuracy. For expression classification, we used $0.67 \cdot \mathrm{F1}+0.33 \cdot \mathrm{Acc}$ as the metric, where F1 denotes the unweighted F1 score for 7 classes, and Acc is the total accuracy. For valence-arousal estimation, we evaluated with the Concordance Correlation Coefficient (CCC).

 \section{Results}
 \subsection{CNN Results (Validation Set)}
 We trained single-task CNN models for all three tasks. Table~\ref{tab: single-CNN} reports the results of the baseline network, after the addition of external data, and after data balancing. We find that both improve the performance of the single-task CNNs for facial AU detection and expression classification by a large margin. The improvement for the  (valence-arousal) regression task is not obvious. However, given that it improved performance on the other tasks and did not degrade valence-arousal, we applied data balancing for the rest of the experiments.
 
\begin{figure}[t] 
\centering
\resizebox{1\columnwidth}!{
\includegraphics[width=\columnwidth]{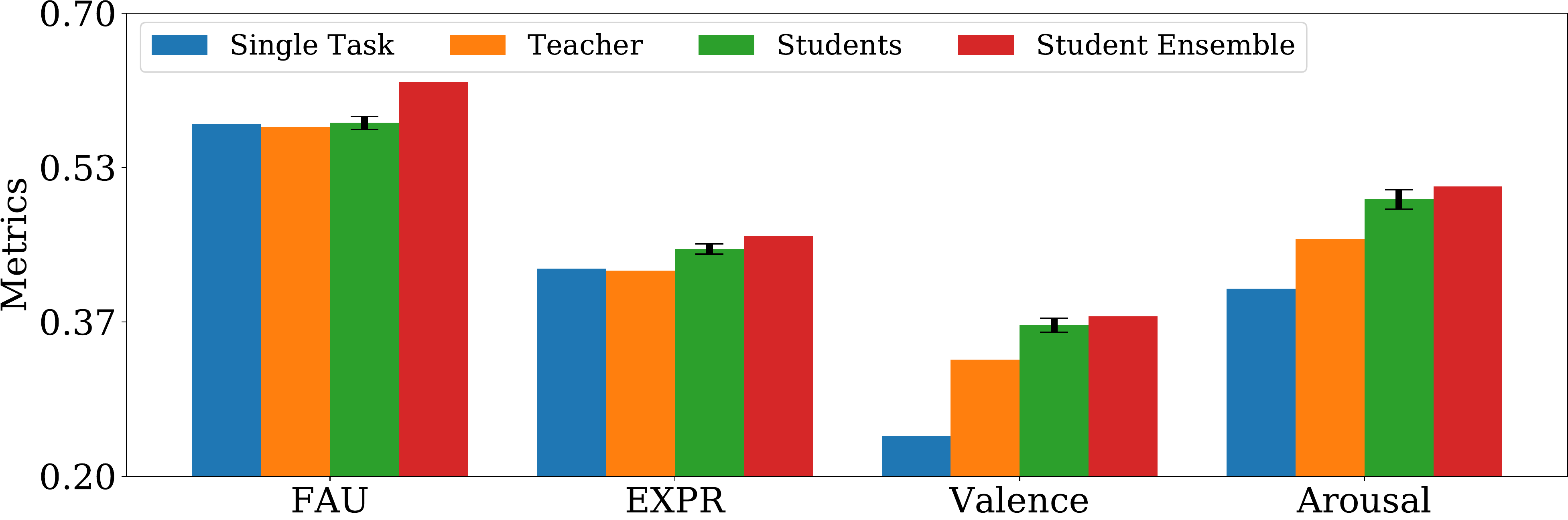} 
}
\caption{Single and multitask CNN performance on the validation set}
\label{fig: multi-CNN}
\end{figure}

 We trained the multitask CNN models using Algorithm~\ref{algorithm1}. Figure~\ref{fig: multi-CNN} compares the performance of the single-task CNN, the performance of the multitask teacher, the average performance of five students, and the performance of the five-student ensemble.
 Comparing multitask teacher and single-task CNN, we find that the multitask teacher model outperforms the single-task model for valence-arousal estimation, but not in facial AU detection and expression recognition. The average performance of the students exceeds that of the teacher on all tasks. We hypothesize that this is due to the supervision by the soft but complete, rather than ground truth but only partial, labels for the students, and the potential "dark knowledge" in the softened probabilities. Ensembling the five students further improves performance.

\begin{table}[tb]
	\centering
	\resizebox{\columnwidth}{!}{
		\begin{tabular}{l|c|c|c|c}
			\hline
			 Methods & facial AU & EXPR & Valence & Arousal \\
			\hline
			 baseline & 0.31 & 0.36 & 0.14 & 0.24 \\
			 \hline
		imbalanced data & 0.5465 & 0.3399 & 0.2406 & \bf 0.4178\\
         \hline
		 balanced data & \bf 0.5802   & \bf 0.4243  & \bf 0.2438 &  0.4027\\
			\bottomrule
		\end{tabular}
		}
	\caption{Single-task CNN performances on the validation set of the Aff-wild2 dataset. Baseline model performances are provided by~\cite{kollias2020analysing}. }
	\label{tab: single-CNN}
\end{table}


 \subsection{CNN-RNN Results (Validation Set)}
 
\begin{figure}[htbp]
  \centering
  \begin{subfigure}[b]{0.9\columnwidth}
    \includegraphics[width=\columnwidth]{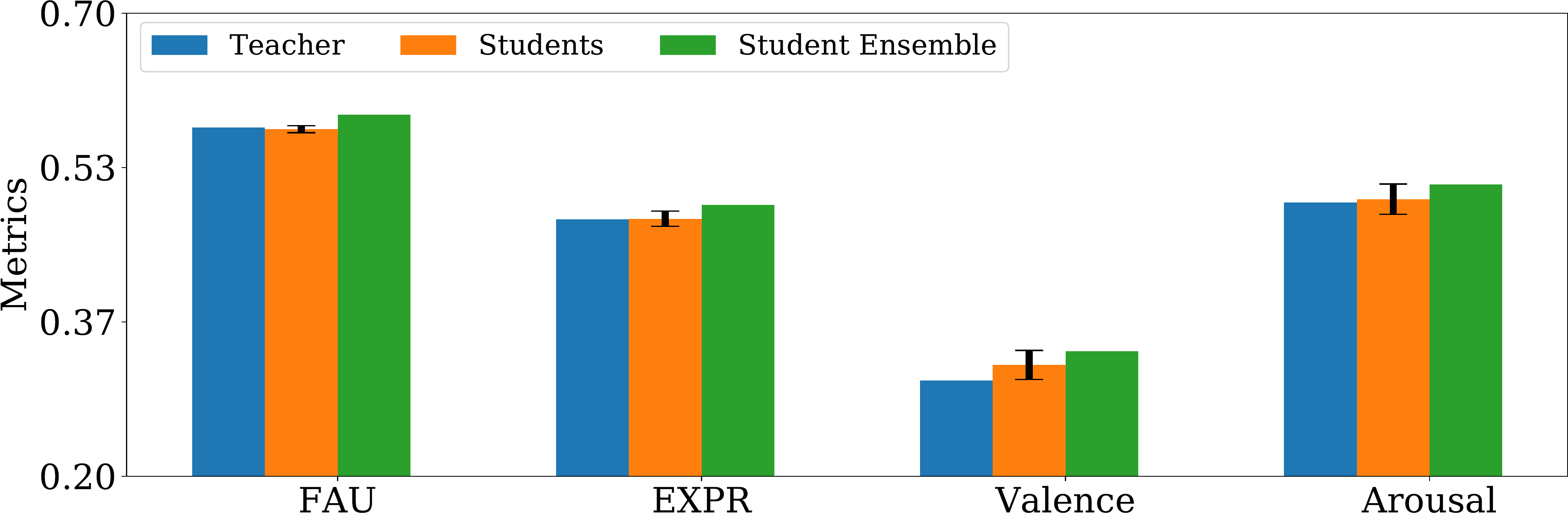}
    \caption{Sequence Length 32}
  \end{subfigure}
  \par\vspace{0.3cm}
  \begin{subfigure}[b]{0.9\columnwidth}
    \includegraphics[width=\columnwidth]{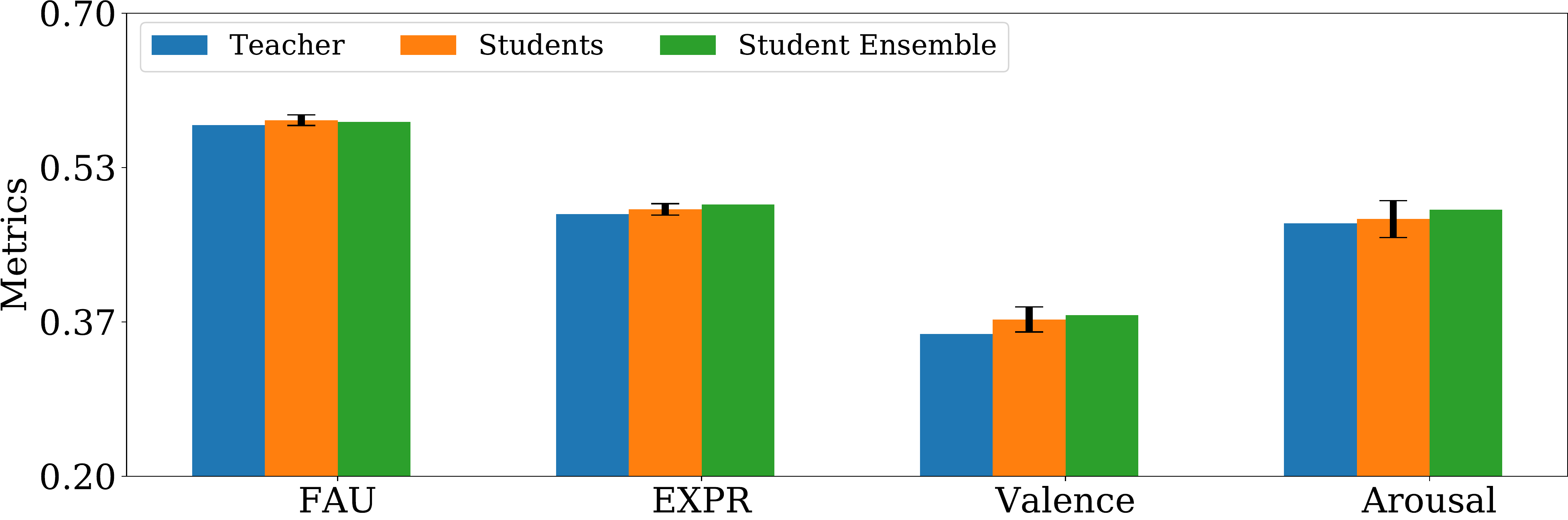}
    \caption{Sequence Length 16}
  \end{subfigure}
  \par\vspace{0.3cm}
  \begin{subfigure}[b]{0.9\columnwidth}
    \includegraphics[width=\columnwidth]{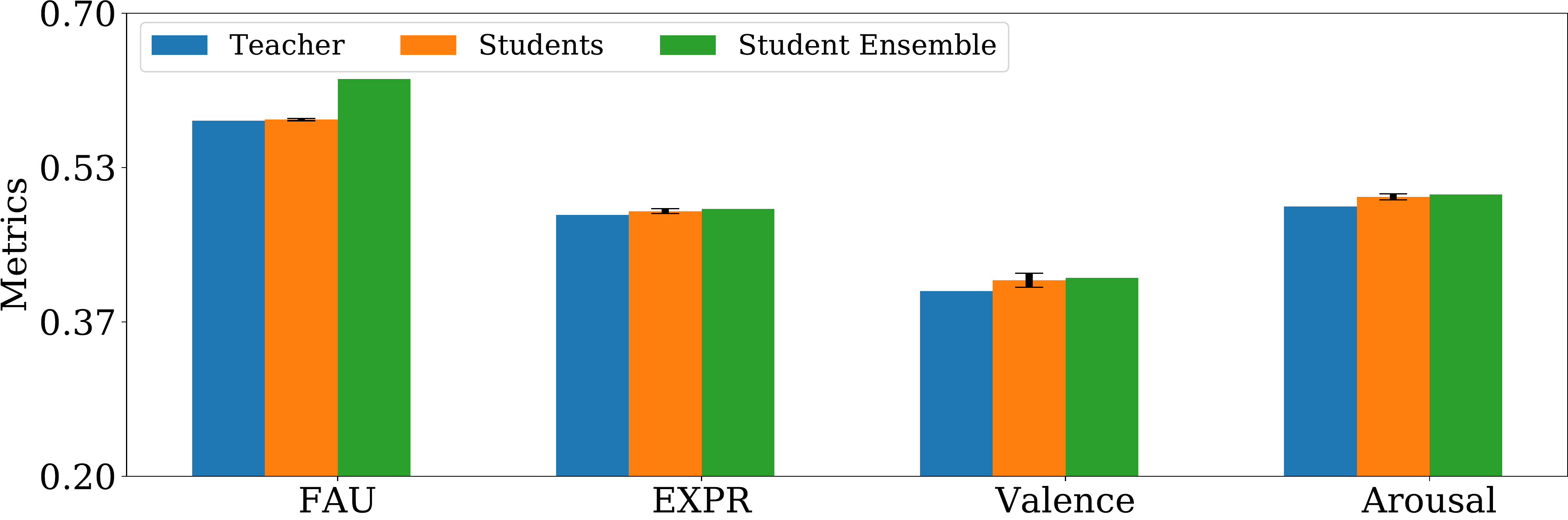}
    \caption{Sequence Length 8}
  \end{subfigure}
  \caption{Multitask CNN-RNN performance on the validation set.}
  \label{fig: multi-CNN_RNN}
\end{figure}

Figure~\ref{fig: multi-CNN_RNN} compares the performance of multitask CNN-RNN teacher, average student and student ensemble models with different sequence lengths during training: 32, 16 and 8 frames. Note that in all models the spatial feature extractor is fixed to that of the  best CNN model to initialize the feature extractor in the CNN-RNN model. Only the parameters of the GRU and final FC layers differ.

On average, the students outperform the teacher model by a small margin for all sequence lengths and tasks, except facial AU detection with sequence length 32. We can further improve their performance by ensembling. The student ensemble gives the best performance in all cases, except for facial AU detection with sequence length 16 where it is surpassed by the average performance of the student models.


\subsection{Results on the Test Set}

Table~\ref{tab: test} reports the performance of several multitask models (teacher, best performing student, and student ensemble) on the test set. The results are consistent with those reported above for the validation set.
First, the student generally outperforms the teacher. Second, the student ensemble generally gives the best performance.  Third, the multitask CNN-RNN significantly outperforms the CNN on all three tasks, emphasizing the importance of temporal information for emotion recognition.

\begin{table}
\small
    \centering
    \resizebox{\columnwidth}{!}{
        \begin{tabular}{l|c|c|c|c|c|c}\hline
        Architecture &\makecell{Sequence\\ Length}& Name  & facial AU & EXPR & Valence & Arousal \\ \hline
       \multirow{3}{*}{CNN} & \multirow{3}{*}{NA} & Teacher & 0.562 &  0.332 &  0.367 & 0.412\\ \cline{3-7}
      & & Student0 &0.565 & 0.372 &0.421 &0.387 \\ \cline{3-7}
         & &  \makecell{Ensemble}& 0.576& 0.386 & 0.429 & 0.414 \\ \hline
        \multirow{5}{*}{CNN-RNN}  & \multirow{3}{*}{32} & Teacher & 0.571 & 0.395 & 0.425 & 0.437\\ \cline{3-7}
        & & Student0&0.587 & 0.400 &0.426 &0.452  \\ \cline{3-7}
        & & \makecell{Ensemble}  &\bf0.607 & 0.405 &\bf 0.440 &\bf  0.454\\ \cline{2-7}
        & \multirow{2}{*}{16} & Teacher  & 0.600 & 0.439 & 0.406 & 0.384\\ \cline{3-7}
        & & \makecell{Ensemble} & 0.598 &\bf0.440 &0.425 &0.424\\ \cline{2-7}
        & \multirow{2}{*}{8} & Teacher & 0.585 & 0.438 &0.400 & 0.420\\ \cline{3-7}
        & & \makecell{Ensemble}  & 0.599 & 0.437 & 0.404& 0.436\\ \bottomrule
        \end{tabular}
        }
    \caption{Performance of our models on the test set. Student0 is the best performing student on the validation set.}
    \label{tab: test}
\end{table}

\section{Conclusion}

This paper has explored data balancing techniques and their application to multitask emotion recognition. We find that data balancing is beneficial for classification, but not necessarily for regression although it does not seem to hurt. We also propose an teacher-student paradigm for learning multitask models in the presence of missing labels. This model is generic, and can be applied to many multitask scenarios aside from emotion recognition, as studied here. Our results show students generally outperform their teacher model on all tasks, and that ensembling students leads to the best performance. We suggest that although the soft yet complete labels provided by the teacher are not necessarily fully reliable, the increased supervision and dark knowledge they provide enabling the students to distill knowledge that enhances generalization. 

\section{ACKNOWLEDGMENTS}

This work was supported in part by the Hong Kong Research Grants Council under grant  number 16244416 and by the Hong Kong  Innovation and Technology Fund under grant number ITS/406/16FP.


\bibliography{FG-2020}
\bibliographystyle{ieee}

\end{document}